\documentclass[a4paper]{article}

\usepackage{INTERSPEECH2019}
\usepackage{enumitem}
\usepackage{tipa}
\usepackage{pgfplots}

\title{On the Use/Misuse of the Term `Phoneme'}
\name{Roger K. Moore, Lucy Skidmore}
\address{Speech and Hearing Research Group, Dept.\ Computer Science, University of Sheffield, UK}
\email{\{r.k.moore,lskidmore1\}@sheffield.ac.uk}

\begin{document}

\maketitle

\begin{abstract}

The term `phoneme' lies at the heart of speech science and technology, and yet it is not clear that the research community fully appreciates its meaning and implications.  In particular, it is suspected that many researchers use the term in a casual sense to refer to the sounds of speech, rather than as a well defined abstract concept.  If true, this means that some sections of the community may be missing an opportunity to understand and exploit the implications of this important psychological phenomenon.  Here we review the correct meaning of the term `phoneme' and report the results of an investigation into its use/misuse in the accepted papers at INTERSPEECH-2018.  It is confirmed that a significant proportion of the community (i) may not be aware of the critical difference between `phonetic' and `phonemic' levels of description, (ii) may not fully understand the significance of `phonemic contrast', and as a consequence, (iii) consistently misuse the term `phoneme'.  These findings are discussed, and recommendations are made as to how this situation might be mitigated.

\end{abstract}

\section{Introduction} \label{sec:INTRO}

The idea that speech is organised around a finite set of `fundamental' sound units is an ancient one, and many languages have exploited this phenomenon in the development of their writing systems \cite{Daniels1996,Sampson2016}.  Of course the study of such sound structures is the primary remit of the speech sciences, specifically the fields of `phonetics' (concerned with the articulatory and acoustic correlates of speech sounds \cite{Ladefoged1962,OConnor1974,Stevens1998}) and `phonology' (concerned with the organisation and usage of speech sounds in any particular language \cite{Jones1973,Bybee2001}).  Likewise, the field of `speech technology' has eagerly embraced the idea that speech may be modelled as the composition of a small set of fundamental units, and such data structures have long been implicated in the algorithms underpinning practical large-vocabulary continuous speech recognition and text-to-speech synthesis systems \cite{Holmes2002,Woodland2002,Gales2007,Taylor2009,Pieraccini2012}.

Of particular interest here is the concept of a `phoneme', commonly defined as ``\emph{the smallest unit of speech that distinguishes one word from another in a particular language}'' \cite{phondef1,phondef2,phondef3,phondef4}.  Such a notion is central to both phonetics and phonology, and it also plays a key role in much of speech technology\footnote{Of course the emergence of `end-to-end' systems raises interesting issues about the role and value of explicit intermediate levels of representation, and these are discussed in Section \ref{sec:MATT}.}.  However, there is a suspicion that the term `phoneme' is often used in a casual non-scientific sense, based on a mistaken belief that it is a concrete acoustic building block rather than an abstract psychological phenomenon.  In particular, it is hypothesised that many speech researchers (i) may not be aware of the critical difference between `phonetic' and `phonemic' levels of description, (ii) do not fully understand the significance of `phonemic contrast', and as a consequence, (iii) consistently misuse the term `phoneme'.  If this is true, then contemporary speech science and technology may be failing to exploit a crucial aspect of spoken language behaviour, and thus be unnecessarily limiting the explanatory power of our models and the capabilities of our technological solutions.
 
This paper reports the results of an investigation into the use/misuse of the term `phoneme', and offers suggestions as to how the situation might be mitigated.  Section \ref{sec:PHON} reviews the definition of the term, Section \ref{sec:STUDY} describes the investigation, and Section \ref{sec:DISC} discusses the implications and makes three key recommendations.  Finally, Section \ref{sec:CONC} concludes with summary of the main findings.

\section{The `Phoneme'} \label{sec:PHON}

\subsection{Background} \label{sec:BACK}

According to the pioneering phonetician Daniel Jones, the idea of the phoneme was recognised from the 1870s, but the term itself was not in general use until the beginning of the 20\textsuperscript{th} century \cite{Jones1973}.  The need for such a term arose because early phoneticians had realised that acoustically distinct speech sounds were \emph{only} perceived as different (by native listeners) \emph{if} they signalled the difference between one word and another (in that language).  Crucially, acoustically distinct speech sounds (in a language) were perceived as the \emph{same} if they did \emph{not} signal the difference between one word and another (in that language).  In other words, the sounds listeners perceive - the `phonemes' - are conditioned on the \emph{meaning} of an utterance, not on a fixed set of acoustic properties.  Daniel Jones thus defined a phoneme as ``\emph{a family of uttered sounds\footnote{Subsequently termed `allophones'.} (segmental elements of speech) in a particular language\footnote{Jones clarified that in referring to `language', he should really use the term `idiolect', i.e.\ language as used by a particular individual.} which count for practical purposes as if they were one and the same}'' \cite[pp.\ 22]{Jones1973}.

This new understanding of the dual physical and psychological nature of speech led to the realisation that any given utterance may be transcribed using \emph{two} levels of description:  the \emph{language-dependent} `phonemic' level (originally referred to as `psychophonic') and the \emph{language-independent} `phonetic' level (originally referred to as `physiophonic').   It also led to the requirement for agreed phonemic and phonetic transcription conventions, the founding of the International Phonetic Association (IPA) in the late 19\textsuperscript{th} century, and the establishment of the international phonetic alphabet \cite{ipa,MacMahon1986}.

As is now well established, the convention is that a \emph{phonemic} transcription consists of IPA symbols between forward slashes, and a \emph{phonetic} transcription consists of IPA symbols between square brackets\footnote{It should be noted that the use of the same IPA symbol set for both phonemic and phonetic transcriptions has long been a potential source of confusion between the two for naive users.}.  For example, the English phrase ``\emph{law and order}'' could be transcribed phonemically as \textipa{/lO: \ae{}nd O:d3/}, but a particular utterance could be transcribed phonetically as \textipa{[lO:r@nO:d@]}, where various phonological processes (e.g.\ elisions, assimilations, epentheses and reductions) explain the relationship between the two \cite{Ashby2005}.

\subsection{Implications} \label{sec:IMP}

There are many implications that arise from the phonemic structure of spoken language.  There is not space here to review the entire field, but two psychological consequences are worthy of mention.  First, individual speech sounds may be `heard'\footnote{I.e.\ `perceived'.} even though they are not present, and second, whole words or phrases may be `heard' even though they are not present!

The first of these is known as the `phoneme restoration effect' \cite{Warren1970}.  Richard Warren showed that if a short section of speech was cut out and replaced by another sound (such as a cough), listeners could not detect that anything was missing; the excised sound was \emph{restored} in the mind of the listener.

The second effect is illustrated nicely by the phonetician Sara Hawkins \cite{Hawkins2003}.  On hearing a verbal enquiry from a family member as to the whereabouts of some mislaid object, the interlocutor might reply with any of the following utterances:
\begin{quote}\vspace{-0.5em}
	\center{
		\textipa{[aI d@Unt n@U]} \\
		\textipa{[aI dUn@U]} \\
		\textipa{[dUn@]} \\
		\textipa{[\~@\~@\~@]}	
	}
\end{quote}

... where the last utterance is barely more than a series of nasal grunts!  Which of these utterances is actually spoken would depend on the communicative context.  Indeed, the example illustrates how speakers and listeners continuously balance the effectiveness of communication against the \emph{effort} required to communicate effectively \cite{Lombard1911} - behaviour that leads to a `contrastive'\footnote{I.e.\ `discriminative'.} (as opposed to signal-based) form of communication \cite{Lindblom1990}.  However, the point here is that the listener perceives \textipa{/aI d@Unt n@U/} (``\emph{I don't know}'') in each case!  Likewise, since it is \emph{top-down} context that facilitates the appropriate perception, the utterance \textipa{[\~@\~@\~@]} might be easily perceived as a completely different sequence of words in a different scenario.

Other significant phenomena include `coarticulation' which, contrary to what most speech technologists think, is not just a local effect, but which can span entire syllables \cite{Ashby2005}.  Indeed, the spread of phonemic information over time (due to asynchronous control of the articulators) means that, almost by definition, there are \emph{no} acoustic boundaries between phonemes (thereby rendering any `beads-on-a-string' assumptions fundamentally flawed \cite{Ostendorf1999}).
	
Of course, all of the above should be familiar to anyone working in speech research.  However \ldots

\section{The Study}  \label{sec:STUDY}

In order to gauge the usage of the term `phoneme' in the broad speech science and technology community, it was decided to analyse the texts of all papers accepted for publication at the most recent INTERSPEECH conference - INTERSPEECH-2018 - which took place in Hyderabad, India in August 2018.  791 papers comprising a total of over 3 million words were analysed, and it was found that 34\% of the papers contained at least one occurrence of the word ``\emph{phoneme}'', with an average of 7.69 occurrences per paper\footnote{One paper contained 88 occurrences of the term `phoneme'!}.

To put these figures into context, Table \ref{tab:IS18} shows the statistics for various other keywords occurring in the INTERSPEECH-2018 accepted papers.  As can be seen, ``\emph{phoneme}'' occurred more frequently than ``\emph{speech synthesis}'', but half as often as ``\emph{speech recognition}''.  Unsurprisingly, ``\emph{speech}'' occurred in all papers, with an average of 44.46 occurrences per paper.

\begin{table}[h]
	\caption{Usage of the terms `phoneme', `speech', `speech recognition' and `speech synthesis' in the 791 accepted papers at INTERSPEECH-2018.  \textbf{Count} refers to the total number of occurrences, \textbf{\#P} and \textbf{\%P} refer to the number and percentage of papers in which the term appears respectively, \textbf{Av.} refers to the average number of occurrences in those papers, and \textbf{Max.} refers to the maximum number of occurrences in any one paper.}
	\label{tab:IS18}
	\centering
	\begin{tabular}{| c | c | c | c | c |}
		\hline
	   	& \emph{phoneme} & \emph{speech} & \emph{speech rec.} & \emph{speech synth.} \\ \hline
	    	\textbf{Count} & 2038 & 35171 & 3210 & 780 \\ \hline
	    	\textbf{\#P} & 265 & 791 & 536 & 152 \\ \hline
	    	\textbf{\%P} & 34\% & 100\% & 68\% & 19\% \\ \hline
		\textbf{Av.} & 7.69 & 44.46 & 5.99 & 5.13 \\ \hline
		\textbf{Max.} & 88 & 196 & 35 & 23 \\ \hline
	\end{tabular}
  \end{table}

\subsection{Historical Comparison} \label{sec:HIST}

For a historical perspective, Table \ref{tab:ICSLP98} shows the statistics for the same keywords occurring in papers accepted for the 5\textsuperscript{th} International Conference on Spoken Language Processing - ICSLP-1998 - which took place in Sydney Australia,  20 years earlier than the Hyderabad INTERSPEECH.  Comparing the data shown in Tables \ref{tab:IS18} and \ref{tab:ICSLP98}, it can be seen that the overall pattern of usage has not changed greatly over the intervening period (apart from the rather surprising observation that the word `speech' was significantly less frequent in ICSLP-1998 - and was even missing from 2\% of the papers!).

\begin{table}[h]
	\caption{Usage of the terms `phoneme', `speech', `speech recognition' and `speech synthesis' in the 831 accepted papers at ICSLP-1998.}
	\label{tab:ICSLP98}
	\centering
	\begin{tabular}{| c | c | c | c | c |}
		\hline
	   	& \emph{phoneme} & \emph{speech} & \emph{speech rec.} & \emph{speech synth.} \\ \hline
	    	\textbf{Count} & 2505 & 19147 & 2251 & 463 \\ \hline
	    	\textbf{\#P} & 290 & 816 & 482 & 140 \\ \hline
	    	\textbf{\%P} & 35\% & 98\% & 58\% & 17\% \\ \hline
		\textbf{Av.} & 8.64 & 23.46 & 4.67 & 3.31 \\ \hline
		\textbf{Max.} & 75 & 121 & 29 & 20 \\ \hline	
	\end{tabular}
  \end{table}

\begin{figure}[h]
	\centering
	\resizebox {\columnwidth} {!} {
	\begin{tikzpicture}
		\begin{axis}[
			ylabel=\bf{No. of papers},
			y label style={at={(axis description cs:0.05,0.5)}},
			xlabel=\bf{No. of occurrences of the term `phoneme'},
			x label style={at={(axis description cs:0.5,0)}},
			enlargelimits=0.05,
			legend style={at={(0.39,0.6)},
			anchor=south west,legend columns=-2},
			ybar interval=0.7,
			]
			\addplot  [fill=white]
				coordinates {(0,526)(1,100)(2,45)(3,28)(4,13)(5,6)(6,10)(7,4)(8,7)(9,3)};
			\addplot  [fill=black]
				coordinates {(0,517)(1,81)(2,22)(3,22)(4,20)(5,14)(6,14)(7,9)(8,8)(9,12)};
			\legend{IS-2018,ICSLP-1998}
		\end{axis}
	\end{tikzpicture}
	}
	\caption{Distribution of the occurrences of the term `phoneme' in the INTERSPEECH-2018 and ICSLP-1998 accepted papers.}
	\label{fig:HIST}
\end{figure}
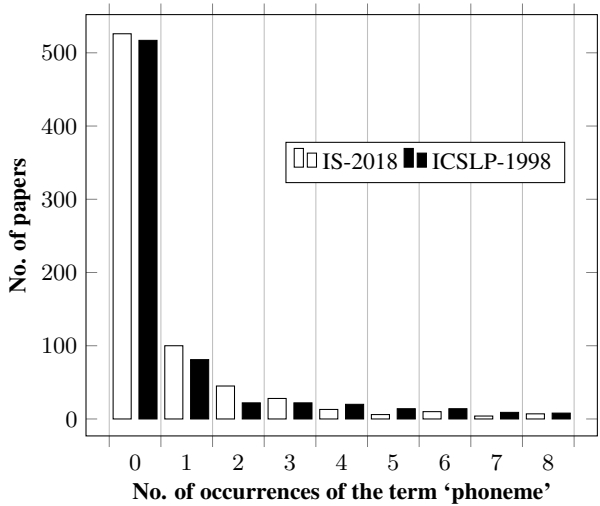

Figure \ref{fig:HIST} shows the distribution of the occurrences of the term `phoneme' in the accepted papers for the two conferences. As can be seen, the distributions are quite similar; the term did not appear at all in 66\% of the INTERSPEECH-18 papers and 65\% of the ICSLP-98 papers.

Both data sets were also analysed to determine the most frequent words to occur immediately before or immediately after the word ``\emph{phoneme}''.  The results are shown in Table \ref{tab:BIGRAM}.  As can be seen, there is a reasonable amount of agreement, with similarly high occurrences of ``\emph{phoneme based}'', ``\emph{phoneme recognition}'' and ``\emph{phoneme sequence}'' in both conferences.  However, what is particularly noticeable is that most of these frequently occurring bigram phrases are more related to speech technology than to speech science.

\begin{table}[h]
	\caption{List of the most frequent words to occur adjacent to the word ``phoneme'' in the accepted papers at the INTERSPEECH-2018 and ICSLP-1998 conferences.}
	\label{tab:BIGRAM}
	\centering
	\begin{tabular}{ c | c }
		\toprule
		\textbf{IS-18} & \textbf{ICSLP-98} \\ \midrule
	     	\emph{based} & \emph{recognition} \\
	    	\emph{sequence} & \emph{similarity} \\
	    	\emph{recognition} & \emph{model} \\
	    	\emph{label} & \emph{sequence} \\
	    	\emph{level} & \emph{based} \\
	    	\emph{conversion} & \emph{boundary} \\
	    	\emph{set} & \emph{duration} \\
	    	\emph{CI} & \emph{string} \\
	    	\emph{classifier} & \emph{class} \\
	    	\emph{classification} & \emph{dependent} \\ \bottomrule
	\end{tabular}
  \end{table}

\subsection{Accuracy of Definitions} \label{sec:DEF}

One potentially clear indicator of the appropriate or inappropriate use of the term `phoneme' is to appraise the accuracy/adequacy of any provided definitions.  However, of the 265 accepted papers in INTERSPEECH-2018 that contained ``\emph{phoneme}'', only six had anything approaching a definition and, of those, none were satisfactory\footnote{Note that citations to individual papers will not be provided in order not to embarrass the author(s).}.  For example, papers contained statements such as ``\emph{Speech signal consists of various basic speech sound units, which are called as phonemes.}'' and ``\emph{\ldots treat any sub-word acoustic unit as a phoneme.}''.  One pseudo-definition was simply incorrect: ``\emph{In phonetics, it is believed that when one pronounces two neighbouring phonemes, there often exists joint frames that can be a very short pause belonging to neither phoneme \ldots}''.  The remaining efforts superficially equated phonemes with ``\emph{sound symbols}'' or with ``\emph{sub words}''.  Interestingly, of these six attempts at a definition, five were in papers presented in speech technology (as opposed to speech science) sessions.

However, what is perhaps most concerning is that 98\% of the INTERSPEECH-2018 papers that used the term `phoneme' did not provide any form of definition or explanation, presumably because the authors assumed that everyone knows what it means.

\subsection{Use and Misuse} \label{sec:UM}

Further analysis of the 265 accepted papers in INTERSPEECH-2018 that contained the term `phoneme' showed that around 40\% used the term in a way that could be construed as misuse.  For example, some authors seemed not to be aware of the crucial difference between phonemic and phonetic levels of description and the associated /\ldots/ versus [\ldots] convention for transcription.  Other authors clearly assumed that there is a fixed relationship between a `phoneme' and its acoustic realisation, and that boundaries between `phonemic segments'  existed and were well defined.  As a consequence, many authors referred to `phonemic segmentation' and the derivation of `phoneme durations'.  One even referred to ``\emph{phoneme chunks}''!

Other examples of misuse include the following:
\begin{itemize}[topsep=0pt, noitemsep, leftmargin=*]
	\item ``\emph{We have 252 phonemes, of which there are 213 Mandarin and 39 English.}'' -- illustrates a fundamental misunderstanding of how phonemes are defined and enumerated.
	\item ``\emph{There are 144 traditional phoneme states in a mono phone system.}'' -- reveals a confusion between levels of representation.
	\item ``\emph{\ldots a set of isolated phonemes extracted from CS [continuous speech] sentences.}'' -- shows a lack of understanding of articulatory dynamics.
	\item ``\emph{\ldots context-dependent phonemes.}'' -- a lack of specificity as to what level of representation is being modelled.
	\item ``\emph{\ldots treating filled pause as a special `phoneme'.} '' -- demonstrating a cavalier application of the term to a non-linguistic event.
	\item ``\emph{Spectral transitions between phonemes \ldots}'' -- false assumption that phonemes are acoustic units.
	\item  ``\emph{There is no clear interpretation of HMM states for emotion recognition as for automatic speech recognition (sub phoneme).}'' -- casual interpretation of different types of representation.
	\item ``\emph{We propose a language-independent phoneme segmentation method.}'' -- shows a lack of understanding of the essential language-specific nature of phonemes.
	\item ``\emph{Even though they all use the same phoneme symbols, each language and accent imposes its own coloring or `twang'.}'' -- overly informal description of phonetic variability.
	\item ``\emph{\ldots modeling only the correct pronunciation of each individual phoneme.}'' -- gross assumption about phonetic variation.
	\item ``\emph{we consider the AUs [acoustic units] as phonemes }'' -- false assumption that phonemes are acoustic units.
	\item ``\emph{\ldots fMLLR normalized features which are speaker independent phoneme representations.}'': false assumption that phonemes are acoustic units.
	\item ``\emph{\ldots If a phoneme lasts for more than 5ms \ldots}'' --  false assumption that phonemes are acoustic units.
	\item ``\emph{\ldots below the minimum duration of a phoneme (30 ms) are considered as spurious regions.}'' -- ignoring the realities of speech production.
	\item ``\emph{Diphthongs and triphtongs [sic] are split into their constituent phones to reduce the number, and enforce sharing, of phonemes.}'' -- failure to appreciate that diphthongs and triphthongs may be phonemes themselves.
	\item ``\emph{\ldots universal phoneme mapping \ldots}'' -- gross assumption about the relationship between the sounds in different languages.
	\item ``\emph{At a local, temporally constrained level, we observe concrete linguistic events (phonemes) \ldots}'' -- misunderstanding about the abstract psychological nature of phonemes.
	\item ``\emph{\ldots trained with context-independent phoneme states as targets.}'' -- false assumption that phonemes are acoustic units.
\end{itemize}

Of course, it should be acknowledged that around 60\% of the 265 accepted papers in INTERSPEECH-2018 that contained the term `phoneme' did \emph{not} misuse the term in an inappropriate manner.  In particular, it was noticeable that authors in the areas of L2 learning and low-resource languages were considerably more precise in their usage.  However, most other mentions were single occurrences where the term `phoneme' was used in a casual/generic sense, e.g.\ to refer to a category label in a classifier.

\subsection{Science vs.\ Technology} \label{sec:ST}

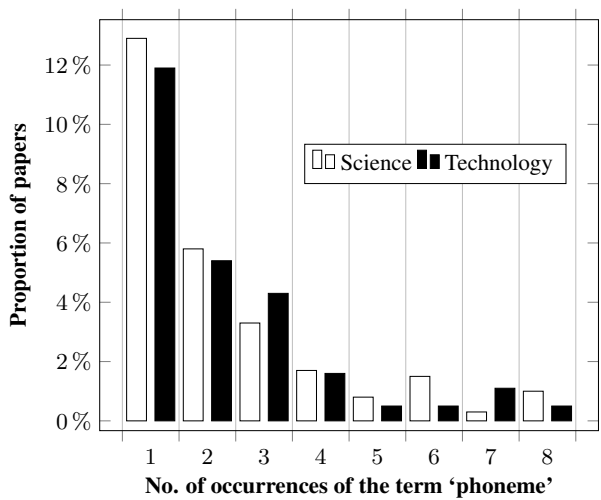
\begin{figure}[h]
	\centering
	\resizebox {\columnwidth} {!} {
	\begin{tikzpicture}
		\begin{axis}[
			ylabel=\bf{Proportion of papers},
			y label style={at={(axis description cs:0.02,0.5)}},
			yticklabel=\pgfmathprintnumber{\tick}\,$\%$,
			xlabel=\bf{No. of occurrences of the term `phoneme'},
			x label style={at={(axis description cs:0.5,0)}},
			enlargelimits=0.05,
			legend style={at={(0.41,0.6)},
			anchor=south west,legend columns=-2},
			ybar interval=0.7,
			]
			\addplot  [fill=white]
				coordinates {(1,12.9)(2,5.8)(3,3.3)(4,1.7)(5,0.8)(6,1.5)(7,0.3)(8,1.0)(9,0.3)};
			\addplot  [fill=black]
				coordinates {(1,11.9)(2,5.4)(3,4.3)(4,1.6)(5,0.5)(6,0.5)(7,1.1)(8,0.5)(9,0.5)};
			\legend{Science,Technology}
		\end{axis}
	\end{tikzpicture}
	}
	\caption{Distribution of the occurrences of the term `phoneme' in the INTERSPEECH-2018 accepted papers in the speech science and speech technology categories.}
	\label{fig:ST}
\end{figure}

As mentioned above, it is conceivable that the term `phoneme' is used differently in in different parts of the research community.  In order to test this, all of the accepted papers in INTERSPEECH-2018 were categorised according to whether they fell into the speech science or speech technology areas.  This resulted in 185 papers tagged as `science' and 606 papers tagged as `technology'.  Figure \ref{fig:ST} shows the distribution of the occurrences of the term `phoneme' based on this categorisation.  As might be expected, there is evidence that the term `phoneme' occurs slightly more frequently in the speech science papers.  However, it was found that the term did not appear at all in 67\% of the speech science papers and 66\% of the speech technology papers - remarkably similar proportions.

Returning to the discussion of use/misuse in Section \ref{sec:UM}, it turned out that, of the papers mentioning the term `phoneme' in potentially inappropriate ways, 25\% were categorised as `science' and 75\% as `technology'.  However, since there were approximately three times as many speech technology papers than speech science papers, it seems that the accuracy associated with using the term is more or less the same in the two sections of the community; a somewhat surprising result.

\section{Discussion and Recommendations} \label{sec:DISC}

The results of this investigation clearly demonstrate that, although the term `phoneme' is used quite frequently by the speech science and technology community, it is often deployed in a casual informal manner without considering its deeper formal implications.  This means that, in many cases, the term `phoneme' could have been substituted by `phone' with no loss of meaning.  Of course, part of the reason for this situation is that many language resources are supplied with predefined so-called `phonemic' labels, often using their own non-IPA symbol sets.  Clearly, this encourages users of such resources not to question the nature and significance of such labels.  In particular, using forced-alignment with such labels reinforces the superficial `beads-on-a-string' view of speech (c.f.\ Section \ref{sec:IMP}).

\subsection{Why it Matters} \label{sec:MATT}

A reader might be forgiven for asking whether the issues raised in this paper have any great significance for future research in speech science and technology.  Indeed, it is acknowledged that there is discomfort within some sections of the community about the validity and usefulness of the `phoneme' as a theoretical construct \cite{Sampson1974,Port2005}, and some proponents of so-called `end-to-end' systems actively reject the concept \cite{Shafaei-Bajestan2018}.  Nevertheless, the opinion of the authors is that there is a small but not insignificant risk that future research may fail to exploit an important aspect of spoken language behaviour, and thus unnecessarily limit the explanatory power of the derived models and the capabilities of technological solutions.

Using the term `phoneme' correctly is certainly a small price to pay to avoid such outcomes, and may even lead to deeper and more valuable insights into the structure and behaviour of spoken language - especially when coupled with contemporary ideas in deep learning \cite{LeCun2015}, such as `generative adversarial networks' \cite{Goodfellow2014} and `attention models' \cite{Bahdanau2016}.

\subsection{Recommendations} \label{sec:REC}

Based on the observations reported in this paper, it is possible to make three key recommendations \ldots
\begin{enumerate}[topsep=0pt, noitemsep, leftmargin=*]
	\item \textbf{Researchers} should avoid the term `phoneme' unless they are certain of its meaning.  In particular, the term `phone' should be used to describe a generic speech sound, and the term `phoneme' should be reserved to refer to the abstract family of sounds that serve to distinguish one word from another in a particular language.
	\item \textbf{Teachers/supervisors} should ensure that newcomers to the field of speech science/technology are fully briefed on the critical difference between `phonetic' and `phonemic' levels of description, the significance of `phonemic contrast', and the correct usage of the term `phoneme' \cite[pp.\ 206]{Gibbon1997}.
	\item \textbf{Community associations} (such as ISCA and IEEE) should take steps to ensure that their members are aware of the importance of  using the term `phoneme' correctly.
\end{enumerate}

\section{Summary and Conclusion} \label{sec:CONC}

The investigation reported in this paper has confirmed the hypothesis that a significant proportion of the community (i) may not be aware of the critical difference between `phonetic' and `phonemic' levels of description, (ii) may not fully understand the significance of `phonemic contrast', and as a consequence, (iii) consistently misuse the term `phoneme'.  Three key recommendations are made that aim to mitigate the situation.

\newpage

\bibliographystyle{IEEEtran}

\bibliography{phoneme}

\end{document}